\documentclass[10pt,twocolumn,letterpaper]{article}

\usepackage{cvpr} 

\usepackage[accsupp]{axessibility}  

\usepackage{graphicx,amsmath,amssymb,booktabs,tabulary,multirow,array,float}
\usepackage{color}
\newcolumntype{I}{!{\vrule width 3pt}}
\newlength\savedwidth

\newlength\savewidth
\newcommand\shline{\noalign{\global\savewidth\arrayrulewidth
 \global\arrayrulewidth 1pt}%
 \hline
\noalign{\global\arrayrulewidth\savewidth}}

\def\x{\times}
\newcolumntype{x}[1]{>{\centering\arraybackslash}p{#1pt}}

\newcommand{\tablestyle}[2]{\setlength{\tabcolsep}{#1}\renewcommand{\arraystretch}{#2}\centering\footnotesize}

\usepackage[pagebackref,breaklinks,colorlinks]{hyperref}
\usepackage{appendix}
\usepackage[capitalize]{cleveref}
\crefname{section}{Sec.}{Secs.}
\Crefname{section}{Section}{Sections}
\Crefname{table}{Table}{Tables}
\crefname{table}{Tab.}{Tabs.}

\def\cvprPaperID{105} 
\def\confName{CVPR}
\def\confYear{2023}

\begin{document}

\title{You Only Segment Once: Towards Real-Time Panoptic Segmentation}

\author{
Jie Hu, Linyan Huang, Tianhe Ren, Shengchuan Zhang, Rongrong Ji, and Liujuan Cao\thanks{Corresponding author}\\
Key Laboratory of Multimedia Trusted Perception and Efficient Computing, \\Ministry of Education of China, Xiamen University
}

\maketitle

\begin{abstract}
In this paper, we propose YOSO, a real-time panoptic segmentation framework.
YOSO predicts masks via dynamic convolutions between panoptic kernels and image feature maps, in which you only need to segment once for both instance and semantic segmentation tasks.
To reduce the computational overhead, we design a feature pyramid aggregator for the feature map extraction, and a separable dynamic decoder for the panoptic kernel generation.
The aggregator re-parameterizes interpolation-first modules in a convolution-first way, which significantly speeds up the pipeline without any additional costs.
The decoder performs multi-head cross-attention via separable dynamic convolution for better efficiency and accuracy.
To the best of our knowledge, YOSO is the first real-time panoptic segmentation framework that delivers competitive performance compared to state-of-the-art models.
Specifically, YOSO achieves 46.4 PQ, 45.6 FPS on COCO; 52.5 PQ, 22.6 FPS on Cityscapes; 38.0 PQ, 35.4 FPS on ADE20K; and 34.1 PQ, 7.1 FPS on Mapillary Vistas.
Code is available at \url{https://github.com/hujiecpp/YOSO}.
\end{abstract}

\section{Introduction}
\begin{figure*}
\centering
\includegraphics[width=0.98\linewidth]{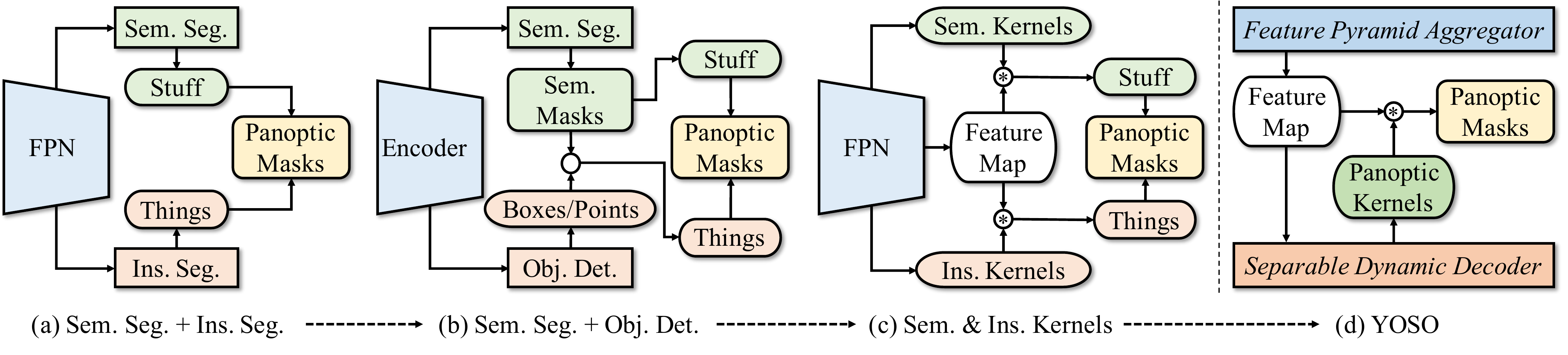} \vspace{-2.4mm}
\caption{
\textbf{Towards real-time panoptic segmentation.}
(a) Semantic and instance segmentation are performed using shared FPN but separated task branches (\eg, in PanopticFPN~\cite{kirillov2019panoptic} and UPSNet~\cite{xiong2019upsnet}).
(b) Semantic segmentation generates masks for all categories, and instance recognition is achieved by object detection using boxes or points (\eg, in RealTimePan~\cite{hou2020real} and PanopticDeepLab~\cite{cheng2020panoptic}).
(c) Kernels for \emph{stuff} and \emph{things} are generated to convolute image feature maps via heavy modules (\eg, in PanopticFCN~\cite{li2021fully}, K-Net~\cite{zhang2021k}, and MaskFormer~\cite{cheng2021per,cheng2021masked}).
(d) YOSO employs an efficient feature pyramid aggregator and a lightweight separable dynamic decoder to produce image feature maps and panoptic kernels.
The figures do not include input images and backbone for concision.
}\label{fig1}
\end{figure*}

Panoptic segmentation is a task that involves assigning a semantic label and an instance identity to each pixel of an input image.
The semantic labels are typically classified into two types, \ie, \emph{stuff} including amorphous and uncountable concepts (such as sky and road), and \emph{things} consisting of countable categories (such as persons and cars).
This division of label types naturally separates panoptic segmentation into two sub-tasks: semantic segmentation for \emph{stuff} and instance segmentation for \emph{things}.
Thus, one of the major challenges for achieving real-time panoptic segmentation is the requirement for separate and computationally intensive branches to perform semantic and instance segmentation respectively.
Typically, instance segmentation employs boxes or points to distinguish between different \emph{things}, while semantic segmentation predicts distribution maps over semantic categories for \emph{stuff}.
As shown in Fig.~\ref{fig1}, numerous efforts~\cite{cheng2020panoptic,li2021fully,tian2022instance,hou2020real,hong2021lpsnet,de2020fast} have been made to unify panoptic segmentation pipelines for improved speed and accuracy.
However, achieving real-time panoptic segmentation still remains an open problem.
On the one hand, heavy necks, \eg, the multi-scale feature pyramid network (FPN) used in~\cite{kirillov2019panoptic,xiong2019upsnet}, and heads, \eg, the Transformer decoder used in~\cite{zhang2021k,cheng2021masked}, are required to ensure accuracy, making real-time processing unfeasible.
On the other hand, reducing the model size~\cite{de2020fast,hong2021lpsnet,hou2020real} leads to a decrease in model generalization.
Therefore, developing a real-time panoptic segmentation framework that delivers competitive accuracy is challenging yet highly desirable.

In this paper, we present YOSO, a real-time panoptic segmentation framework.
YOSO predicts panoptic kernels to convolute image feature maps, with which you only need to segment once for the masks of background \emph{stuff} and foreground \emph{things}.
To make the process lightweight, we design a feature pyramid aggregator for extracting image feature maps, and a separable dynamic decoder for generating panoptic kernels.
In the aggregator, we propose convolution-first aggregation (CFA) to re-parameterize the interpolation-first aggregation (IFA), resulting in an approximately 2.6$\times$ speedup in GPU latency without compromising performance.
Specifically, we demonstrate that the order, \ie, interpolation-first or convolution-first, of applying bilinear interpolation and 1$\times$1 convolution (w/o bias) does not affect results, but the convolution-first way provides a considerable speedup to the pipeline.
In the decoder, we propose separable dynamic convolution attention (SDCA) to perform multi-head cross-attention in a weight-sharing way.
SDCA achieves better accuracy (+1.0 PQ) and higher efficiency (approximately 1.2$\times$ faster GPU latency) than traditional multi-head cross-attention.

In general, YOSO has three notable advantages.
First, CFA reduces computational burden without re-training the model or compromising performance.
CFA can be adapted to any task that uses the combination of bilinear interpolation and 1$\times$1 convolution operations.
Second, SDCA performs multi-head cross-attention with better accuracy and efficiency.
Third, YOSO runs faster and has competitive accuracy compared to state-of-the-art panoptic segmentation models, and its generalization is validated on four popular datasets:
COCO (46.4 PQ, 45.6 FPS), Cityscapes (52.5 PQ, 22.6 FPS), ADE20K (38.0 PQ, 35.4 FPS), and Mapillary Vistas (34.1 PQ, 7.1 FPS).

\section{Related Work}
\textbf{Real-Time Panoptic Segmentation.} Panoptic segmentation aims to jointly perform semantic and instance segmentation, where each pixel in an input image is assigned both a semantic label and a unique instance identity.
Many studies have been conducted for fast panoptic segmentation~\cite{hong2021lpsnet,hou2020real,li2021fully,xiong2019upsnet,de2020fast,cheng2020panoptic,petrovai2020real,chen2020panonet,petrovai2022fast,diaz2021yolo,mohan2021efficientps,yang2019deeperlab,tian2022instance}.
For instance, UPSNet~\cite{xiong2019upsnet} utilizes a deformable convolution based semantic segmentation head and a Mask R-CNN~\cite{he2017mask} style instance segmentation head.
FPSNet~\cite{de2020fast} proposes a fast architecture for panoptic segmentation, avoiding instance mask prediction and merging outputs via soft attention masks.
Recently, PanopticDeepLab~\cite{cheng2020panoptic}, LPSNet~\cite{hong2021lpsnet}, and RealTimePan~\cite{hou2020real} generate semantic masks for all categories first, then locate masks of instances via boxes or points, enabling efficient object segmentation.
Meanwhile, PanopticFCN~\cite{li2021fully}, K-Net~\cite{zhang2021k}, and MaskFormer~\cite{cheng2021per,cheng2021masked} attempt to predict masks for both \emph{things} and \emph{stuff} simultaneously through dynamic convolutions.
Despite significant advances in this field, achieving real-time panoptic segmentation remains an open problem.
In this paper, YOSO enables real-time panoptic segmentation with competitive accuracy by utilizing the proposed feature pyramid aggregator and separable dynamic decoder. 

\textbf{Real-Time Instance Segmentation.} Instance segmentation aims to predict masks and classes for each instance in an image.
To achieve real-time instance segmentation, various approaches have been proposed in recent literature.
YOLACT~\cite{bolya2019yolact,bolya2020yolact++} proposes to multiply the predicted mask coefficients with prototype masks, and SipMask~\cite{cao2020sipmask} utilizes spatial mask coefficients for more accurate segmentation.
CenterMask~\cite{lee2020centermask} employs an efficient anchor-free framework, and DeepSnake~\cite{peng2020deep} explores the use of object contours for fast segmentation of instances.
OrienMask~\cite{du2021real} designs discriminative orientation maps that recover the masks without additional foreground segmentation, and SOLO~\cite{wang2020solo,wang2020solov2} segments objects by locations, with a decoupled branch to speed up the framework.
Recently, SparseInst~\cite{cheng2022sparse} introduced a sparse set of instance activation maps that highlight informative regions for each object in an image, constructing a real-time instance segmentation framework.
As complex operations are required to distinguish different \emph{things}, solving instance segmentation efficiently is the key to real-time panoptic segmentation.
In this paper, YOSO predicts unified panoptic kernels for \emph{stuff} and \emph{things}.
Implementing bipartite matching loss~\cite{carion2020end} for fast discrimination of different \emph{things}, YOSO avoids time-consuming object localization operations such as RoIAlign~\cite{he2017mask} and post-processes such as non-maximum suppression.
The output masks naturally represent independent instances for the categories of \emph{things}.
Moreover, the experimental results in our supplementary material show that YOSO can also achieve competitive performance on real-time instance segmentation.

\textbf{Real-Time Semantic Segmentation.} Semantic segmentation aims to predict pixel-wise categories for input images.
In recent years, many approaches have been developed to enable real-time semantic segmentation.
For example, E-Net~\cite{paszke2016enet} proposes a lightweight architecture for high-speed segmentation, and SegNet~\cite{badrinarayanan2017segnet} combines a small network architecture with skip connections to achieve fast segmentation. 
ICNet~\cite{zhao2018icnet} uses an image cascade algorithm to speed up the pipeline, and ESPNet~\cite{mehta2018espnet,mehta2019espnetv2} introduces an efficient spatial pyramid dilated convolution.
Additionally, BiSeNet~\cite{yu2021bisenet,yu2021bisenet} separates spatial details and categorical semantics to enable both high accuracy and high efficiency in semantic segmentation.
More recently, SegFormer~\cite{xie2021segformer} employs Transformers with a lightweight multi-layer perceptron decoder for fast semantic segmentation.
In contrast to traditional methods that predict distribution maps over classes for semantic masks, YOSO predicts kernels with their corresponding categories for segmentation.
This enables an efficient way to jointly solve semantic and instance segmentation for panoptic segmentation.
\begin{figure}
\centering
\includegraphics[width=0.98\linewidth]{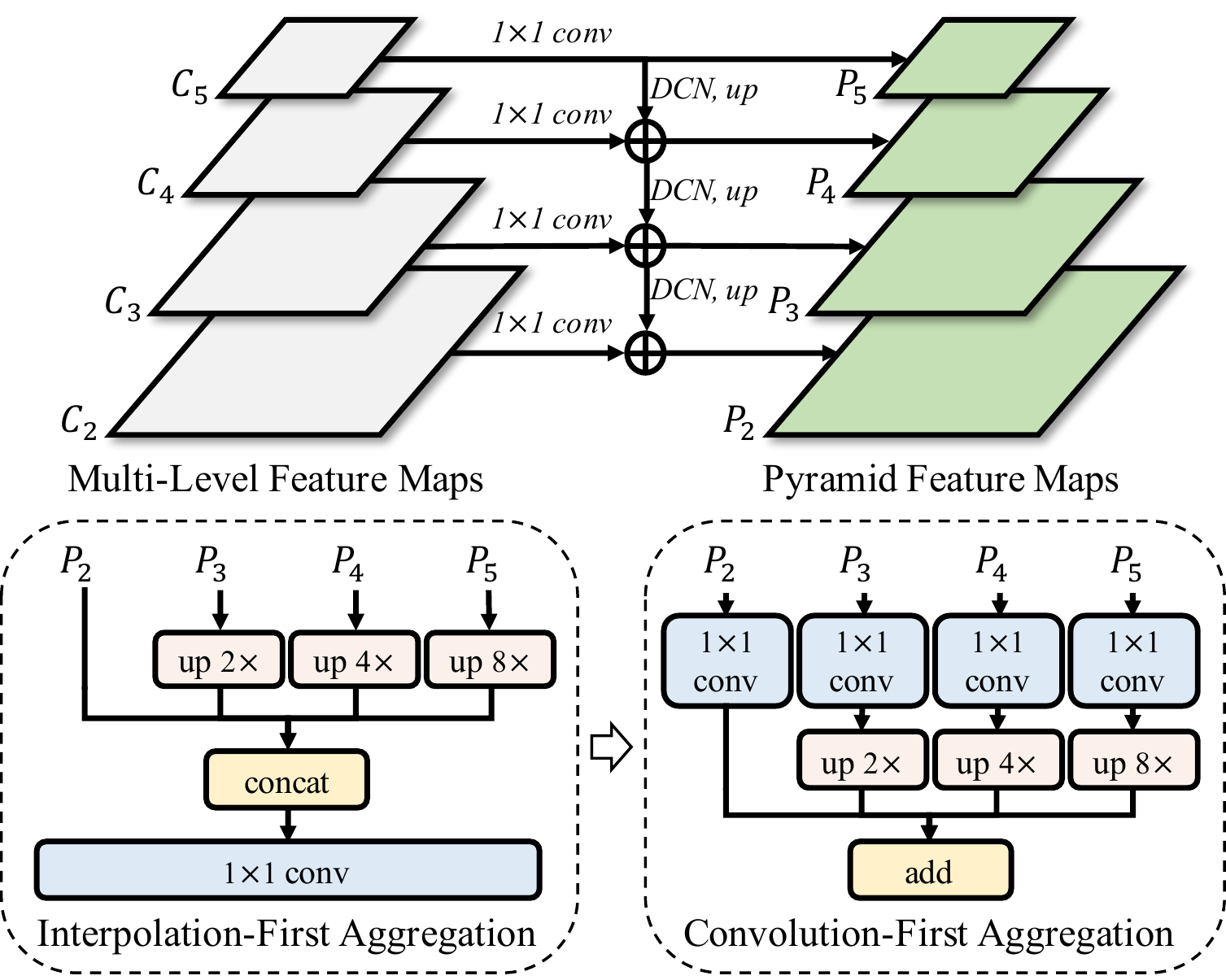} \vspace{-1.0mm}
\caption{
\textbf{Feature pyramid aggregator.}
Multi-level feature maps are fused to generate pyramid features via 1$\times$1 convolutional layers and DCNs.
Then, the pyramid feature maps are aggregated via CFA that can re-parameterize IFA for efficiency.
}\label{fig2}
\end{figure}
\section{Method}
\subsection{Task Formulation}
\textbf{Unified Panoptic Segmentation.} Panoptic segmentation maps each pixel of an image to a semantic class and an instance identity.
We propose a unified approach that considers the set of foreground and background classes as a single entity.
To achieve this, we aim to predict $n$ binary masks $\boldsymbol{M}\in\mathbb{B}^{n\times h\times w}$ with class probabilities $\boldsymbol{L}\in\mathbb{R}^{n\times l}$ for an input image, where $(h, w)$ denotes the mask resolution and $l$ is the total number of categories.
During training, the predictions are matched with the corresponding ground truth labels using the set prediction loss~\cite{carion2020end,cheng2021masked}.
At test time, the segmentation results are merged in terms of the foreground (\ie, \emph{things}) and background (\ie, \emph{stuff}) classes.
Concretely, the masks that correspond to the same background class are merged via union operation; the masks with foreground classes are treated as independent instances; if a pixel belongs to multiple classes, the class with the highest probability is assigned to that pixel.

\textbf{YOSO Framework.}
As shown in Fig.\ref{fig1}, YOSO is a compact framework designed for real-time panoptic segmentation, which consists of a feature pyramid aggregator and a separable dynamic decoder.
The backbone network, such as ResNet\cite{he2016deep}, extracts multi-level feature maps from input images.
The feature pyramid aggregator compresses and aggregates the multi-level feature maps into single-level.
The separable dynamic decoder then generates panoptic kernels with the single-level feature maps for both mask prediction and classification.

\subsection{Feature Pyramid Aggregator}
\textbf{Deformable Feature Pyramid.} Given the multi-level feature maps $\boldsymbol{C}_2,\boldsymbol{C}_3,\boldsymbol{C}_4,\boldsymbol{C}_5$ extracted by the backbone, we utilize the deformable feature pyramid network~\cite{lin2017feature,dai2017deformable} to enhance the feature maps from different scales.
As depicted in Fig.~\ref{fig2}, we first apply 1$\times$1 convolutional layers to compress the channels of the multi-level feature maps.
Then, the feature maps from $\boldsymbol{C}_3$ to $\boldsymbol{C}_5$ are fed into deformable convolutional networks (DCNs) and upsampled level-by-level, yielding the pyramid feature maps $\boldsymbol{P}_2\in\mathbb{R}^{c_2\times h\times w}$, $\boldsymbol{P}_3\in\mathbb{R}^{c_3\times h/2\times w/2}$, $\boldsymbol{P}_4\in\mathbb{R}^{c_4\times h/4\times w/4}$, and $\boldsymbol{P}_5\in\mathbb{R}^{c_5\times h/8\times w/8}$, respectively, where $c_2,c_3,c_4,c_5$ denote the channel dimensions, and $h, w$ represent the 1/4 scale of the input images.
After obtaining the pyramid feature maps, we explore two aggregation methods, namely interpolation-first aggregation (IFA) and convolution-first aggregation (CFA), to merge the multi-level feature maps.

\textbf{Interpolation/Convolution-First Aggregation.} In IFA, the pyramid feature maps are first upsampled to the scale of $h\times w$ via bilinear interpolation.
Then, the feature maps are concatenated and fused using a 1$\times$1 convolutional layer.
In CFA, the pyramid feature maps are first fed to different 1$\times$1 convolutional layers.
Then, the feature maps are bilinearly interpolated to the scale of $h\times w$ and summed.

\textit{\textbf{Observation I:} The output of IFA is exactly equal to that of CFA when using 1$\times$1 convolution without bias.}

This observation is attributed to the homogeneity and additivity properties of the bilinear interpolation function $f(\cdot)$, where $f(\sum_i{w^i\boldsymbol{v}_{x,y}^i})=\sum_i{w^if(\boldsymbol{v}_{x,y}^i)}$ for the constant $w^i$ and the value vector $\boldsymbol{v}_{x,y}^i$ from the four positions $(x_1,y_1), (x_1,y_2), (x_2,y_1), (x_2,y_2)$ of the $i$-th feature map.
Specifically, the bilinear interpolation estimates the value at $(x_0,y_0)$ using the four positions by:
\begin{equation}
\begin{split}
\label{eq1}
f\bigl(\sum\nolimits_{i}w^{i}&\boldsymbol{v}_{x,y}^i\bigr)
=\frac{1}{(x_2-x_1)(y_2-y_1)}
\begin{bmatrix}
 x_2-x_0\\x_0-x_1
\end{bmatrix}^\top \\
&\begin{bmatrix}
 \sum_{i}w^{i}v_{x_1,y_1}^{i},\ \sum_{i}w^iv_{x_1,y_2}^i\\
 \sum_{i}w^{i}v_{x_2,y_1}^{i},\ \sum_{i}w^iv_{x_2,y_2}^i
\end{bmatrix}
\begin{bmatrix}
 y_2-y_0\\
 y_0-y_1
\end{bmatrix}\\
&=\sum\nolimits_{i}{w^if(\boldsymbol{v}_{x,y}^i)},
\end{split}
\end{equation}
where $w^i$ can be interpreted as a 1$\times$1 kernel that convolutes the values of the original positions in the feature maps.
Eq.~\ref{eq1} implies that applying 1$\times$1 convolution (without bias) before or after bilinear interpolation does not affect the final results.
Hence, it can be inferred that when using 1$\times$1 convolution without bias in the aggregators, IFA and CFA produce identical outputs.
\textit{\textbf{Observation II:} CFA requires significantly fewer floating point operations (FLOPs) than IFA.}

The reduction ratio of FLOPs between IFA and CFA is:
\begin{equation}
\begin{split}
\label{eq2}
\frac{4(c_5+c_4+c_3)hw + (c_5+c_4+c_3+c_2)dhw}{(\frac{c_5}{64}+\frac{c_4}{16}+\frac{c_3}{4}+c_2)d h w + 12 d h w + 3 d h w}>1,
\end{split}
\end{equation}
where $d$ is the channel dimension of output feature maps.
In the numerator of Eq.~\ref{eq2} (\ie, the FLOPs of IFA), the first term represents the number of FLOPs used in bilinear interpolation, and the second term represents the number of FLOPs used in 1$\times$1 convolution.
In the denominator of Eq.~\ref{eq2} (\ie, the FLOPs of CFA), the terms represent the number of FLOPs for 1$\times$1 convolution, bilinear interpolation, and accumulation, respectively.
%
%
%

Given the above two observations, we adopt CFA in the proposed feature pyramid aggregator.
It is noteworthy that the learned weights of the 1$\times$1 convolutional layer in IFA can be readily re-parameterized to CFA by dividing the weights into four 1$\times$1 convolutional layers.
This can accelerate the pipeline without incurring any additional costs.

\subsection{Separable Dynamic Decoder}
\begin{figure}
\centering
\includegraphics[width=0.98\linewidth]{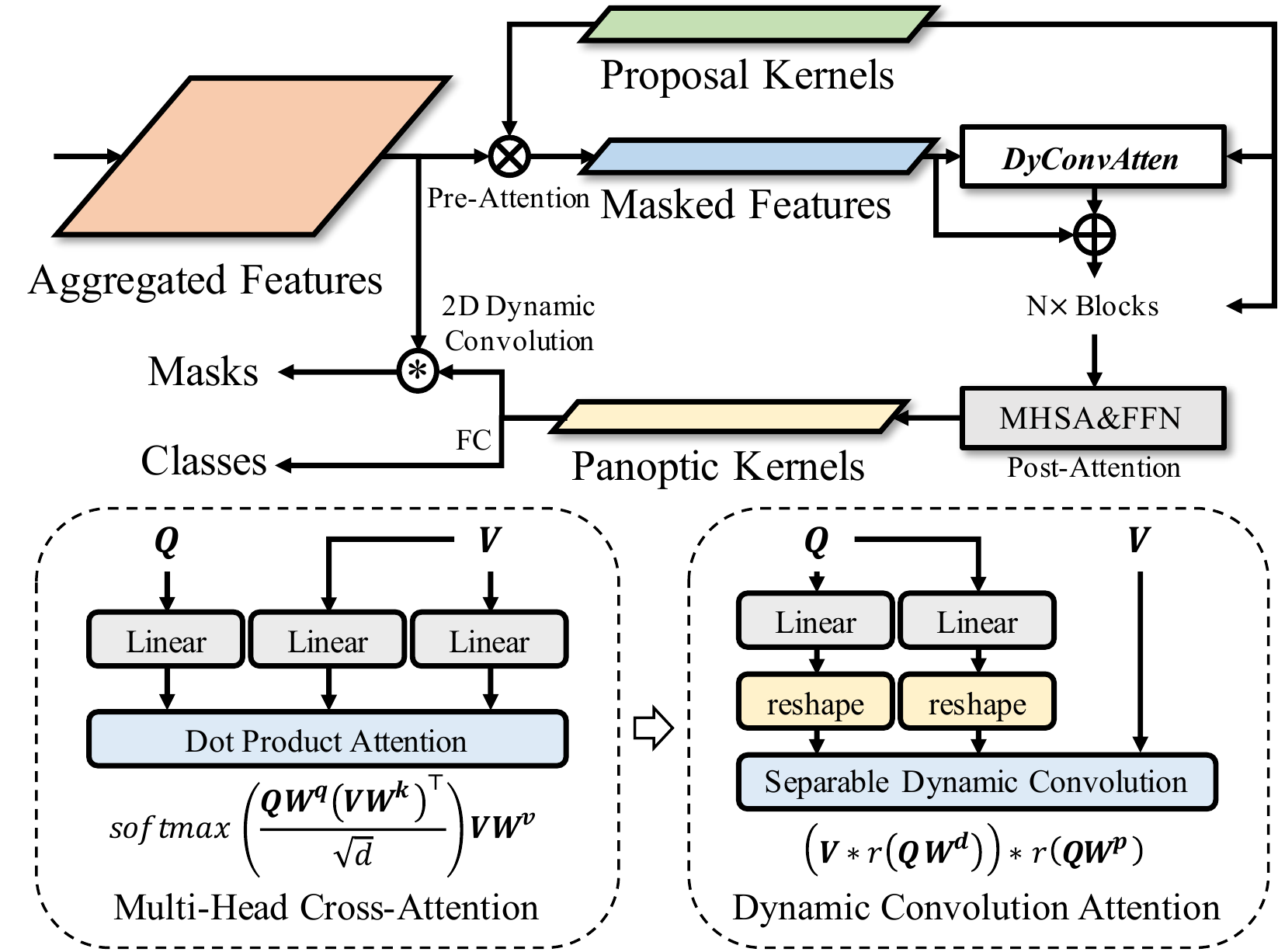}\vspace{-1.0mm}
\caption{
\textbf{Separable dynamic decoder.}
Separable dynamic convolution learns to generate panoptic kernels for 2D dynamic convolutions, which performs multi-head cross-attention in a weight-sharing manner and achieves better accuracy.
}\label{fig3}
\end{figure}
In order to generate accurate kernels for segmentation, previous methods typically relied on dense predictors~\cite{li2021fully} or heavy Transformer decoders~\cite{zhang2021k}.
In contrast, we propose a lightweight kernel generator called the separable dynamic decoder, which speeds up kernel generation while maintaining high accuracy.
The separable dynamic decoder, shown in Fig.~\ref{fig3}, consists of three modules: a pre-attention module, a separable dynamic convolution module, and a post-attention module.
Specifically, the separable dynamic convolution efficiently performs multi-head cross-attention and achieves better accuracy.
We describe each module in detail below.

\textbf{Pre-Attention.}
Inspired by~\cite{zhang2021k}, the pre-attention module selectively extracts key information from the aggregated feature maps to diversify the proposal kernels.
Concretely, the aggregated feature maps $\boldsymbol{S}\in\mathbb{R}^{d\times h\times w}$ are convoluted by the learnable proposal kernels $\boldsymbol{Q}\in\mathbb{R}^{n\times d}$ to produce attention maps $\boldsymbol{A}\in\mathbb{R}^{n\times h\times w}$.
A hard sigmoid function $\sigma(\cdot)$ is applied to the attention maps to activate the input values and discretize the values to either $0$ or $1$ using a threshold of $0.5$.
Then, with the attention maps, the masked features $\boldsymbol{V}\in\mathbb{R}^{n\times d}$ can be obtained by:
\begin{equation}
\begin{split}
\label{eq3}
\boldsymbol{V}=r\bigl(\sigma(\boldsymbol{S}*\boldsymbol{Q})\bigr) {r(\boldsymbol{S})}^\top=r\bigl(\sigma(\boldsymbol{A})\bigr){r(\boldsymbol{S})}^\top,
\end{split}
\end{equation}
where $r(\cdot)$ reshapes $\boldsymbol{A}$ and $\boldsymbol{S}$ to the sizes of $(n, hw)$ and $(d, hw)$, respectively, $*$ denotes 2D convolution operation.

\textbf{Vanilla Dynamic Convolution.}
To increase the model capacity, multi-head cross-attention~\cite{vaswani2017attention} has been widely used, which projects input features into multiple spaces to model their relationships.
For instance, given $\boldsymbol{Q},\boldsymbol{V}\in\mathbb{R}^{n\times d}$ which contains $n$ tokens with the hidden dimension of $d$, a $t$-head cross-attention can be defined as follows:
\begin{equation}
\begin{split}
\label{eq4}
MultiHeadCross&Atten(\boldsymbol{Q},\boldsymbol{V})=\\
&Concat(\boldsymbol{H}_1,\boldsymbol{H}_2,\ldots,\boldsymbol{H}_t)\boldsymbol{W}^o,
\end{split}
\end{equation}
where $\boldsymbol{W}^o\in\mathbb{R}^{d\times d}$ is a linear transformation matrix.
The attention operation inside each head is defined as:
\begin{equation}
\begin{split}
\label{eq5}
\boldsymbol{H}_i&=softmax\bigl(\frac{\boldsymbol{Q}\boldsymbol{W}_i^q(\boldsymbol{V}\boldsymbol{W}_i^k)^\top}{\sqrt{d}}\bigr)(\boldsymbol{V}\boldsymbol{W}_i^v)\\
&=\boldsymbol{K}_i(\boldsymbol{V}\boldsymbol{W}_i^v)=\boldsymbol{K}_i\boldsymbol{V}_i,
\end{split}
\end{equation}
where $\boldsymbol{W}_i^q,\boldsymbol{W}_i^k,\boldsymbol{W}_i^v\in\mathbb{R}^{d\times d/t}$ are the projection matrices, $\boldsymbol{V}_i\in\mathbb{R}^{n\times d/t}$ denotes the projected features, and $\boldsymbol{K}_i\in\mathbb{R}^{n\times n}$ denotes the correlation matrix.

Although increasing the model capacity enhances performance, it also results in high computational burden.
Intuitively, multi-head cross-attention involves three fundamental operations: multi-head projection, cross-token interaction, and cross-dimension interaction.
In Eq.~\ref{eq4}, the cross-dimension interaction, represented by $\boldsymbol{W}^o$, learns to re-weight the importance of every hidden dimension.
In Eq.~\ref{eq5}, the mutli-head projection, represented by $\boldsymbol{V}\boldsymbol{W}_i^v$, maps the hidden dimensions $d$ into $t$ different spaces with the size of $d/t$, and the cross-token interaction, represented by $\boldsymbol{K}_i\boldsymbol{V}_i$, uses the correlation matrix to interact between tokens.
These operations motivated us to perform multi-head cross-attention using 1D convolution to make the process lightweight, which is defined as:
\begin{equation}
\begin{split}
\label{eq6}
Conv_{1d}(\boldsymbol{K},\boldsymbol{V})=\boldsymbol{V}*\boldsymbol{K},
\end{split}
\end{equation}
where $\boldsymbol{K}\in\mathbb{R}^{n\times n\times t}$ denotes the weights of kernels to convolute $\boldsymbol{V}\in\mathbb{R}^{n\times d}$, $*$ denotes 1D convolution operation.
The 1D convolution computes the element of the output $\boldsymbol{O}\in\mathbb{R}^{n\times d}$ at the position $i, j$ by:
\begin{equation}
\begin{split}
\label{eq7}
O_{i, j} = \sum_{p=1}^n{\sum_{q=1}^t{K_{i,p,q}\cdot V_{p,j+q-1}}}.
\end{split}
\end{equation}
Correspondingly, the basic operations of multi-head cross-attention are also performed in 1D convolution in a weight-sharing manner.
For the multi-head projection, the sliding window in 1D convolution densely splits the hidden dimensions into $d$ groups of size $t$, and the $t$ successive hidden dimensions in each group are projected with shared kernels.
For the cross-token interaction, the first accumulation term in Eq.~\ref{eq7} interacts $n$ tokens by $n$ different kernels.
For the cross-dimension interaction, the second accumulation term in Eq.~\ref{eq7} locally incorporates the information from $t$ successive hidden dimensions, instead of using all hidden dimensions globally in multi-head cross-attention.
Furthermore, inspired by~\cite{kitaev2020reformer,choromanski2020rethinking,wang2020linformer,katharopoulos2020transformers,zaheer2020big,FelixWu2019PayLA}, we employ a dynamic approach to generate the kernel $\boldsymbol{K}$ conditioned on $\boldsymbol{Q}$, which introduces the cross-attention mechanism into 1D convolution and defines a dynamic convolution attention as follows:
\begin{equation}
\begin{split}
\label{eq8}
DyConvAtten(\boldsymbol{Q},\boldsymbol{V})=\boldsymbol{V}*r(\boldsymbol{Q}\boldsymbol{W}),
\end{split}
\end{equation}
where $\boldsymbol{W}\in\mathbb{R}^{d\times nt}$ is the projection matrix for generating the dynamic kernels.
%
%

\textbf{Separable Dynamic Convolution.}
The standard convolution can be further decomposed into a depthwise convolution and a pointwise convolution as in~\cite{howard2017mobilenets}.
Following this approach, we propose a separable form for vanilla dynamic convolution as follows:
\begin{equation}
\begin{split}
\label{eq9}
SepDyConvAtten(\boldsymbol{Q}&,\boldsymbol{V})=\\
&\bigl(\boldsymbol{V}*r(\boldsymbol{Q}\boldsymbol{W}^{d})\bigr)*r(\boldsymbol{Q}\boldsymbol{W}^{p}),
\end{split}
\end{equation}
where $\boldsymbol{W}^{d}\in\mathbb{R}^{d\times t}, \boldsymbol{W}^{p}\in\mathbb{R}^{d\times n}$ project $\boldsymbol{Q}$ to generate the kernels for depthwise convolution and pointwise convolution. The kernels are then reshaped to $(n,1,t)$ and $(n,n,1)$, respectively.
In Eq.~\ref{eq9}, the depthwise convolution models the cross-dimension interaction, while the pointwise convolution models the cross-token interaction.
Compared to multi-head cross-attention, the reduction in FLOPs is:
\begin{equation}
\begin{split}
\label{eq10}
\frac{4nd^2+2n^2d}{2ndt+2n^2d}=\frac{2d+n}{k+n}>1.
\end{split}
\end{equation}
In the numerator of Eq.~\ref{eq10} (\ie, the FLOPs of MHCA), the first term denotes the number of FLOPs used in multi-head projection, while the second term represents the number of FLOPs used in cross-attention.
In the denominator of Eq.~\ref{eq10} (\ie, the FLOPs of SDCA), the terms correspond to the number of FLOPs for 1D convolution operation and linear projection, respectively.
%

\textbf{Post-Attention.}
In the post-attention module, we use a multi-head self-attention layer and a feed-forward network to generate the panoptic kernels.
Subsequently, we produce the masks by means of 2D convolution and predict the associated classes using additional feed-forward networks. 
Inspired by~\cite{chen2019hybrid,cai2018cascade}, we utilize the panoptic kernels to update the proposal kernels iteratively for improved accuracy.

\section{Experiments}
\subsection{Datasets and Evaluation Metrics}
\textbf{Datasets.} We evaluated the effectiveness and efficiency of YOSO on four widely used panoptic segmentation datasets: the COCO dataset~\cite{lin2014microsoft}, the Cityscapes dataset~\cite{Cordts2016Cityscapes}, the ADE20K~\cite{BoleiZhou2017ScenePT} dataset, and the Mapillary Vistas~\cite{GerhardNeuhold2017TheMV} dataset.
The COCO dataset gathers images of complex everyday scenes with common objects, containing 80 \textit{things} categories and 53 \textit{stuff} categories in 118k images for training and 5k images for validation.
The Cityscapes dataset contains images of urban street-view scenes, which has 8 \textit{things} categories and 11 \textit{stuff} categories in 2.9k images for training and 0.5k images for validation.
The ADE20K dataset is annotated in an open-vocabulary setting with 50 \textit{things} categories and 100 \textit{stuff} categories, including 20k images for training and 2k images for validation.
The Mapillary Vistas dataset is a large-scale urban street-view dataset with 37 \textit{things} categories and 28 \textit{stuff} categories in 18k and 2k images for training and validation.

\textbf{Evaluation Metrics.} The panoptic segmentation results are assessed using the panoptic quality (PQ) metric~\cite{kirillov2019panoptic1}, which can be further decomposed to the segmentation quality (SQ) and the recognition quality (RQ).
The evaluation outcomes for \emph{stuff} and \emph{things} are represented by the superscripts $s$ and $t$, respectively.
The frame rate, \ie, frames per second (FPS), of the YOSO models is evaluated on a single V100 GPU for the main results and a single 3090 GPU for the ablation study.

\begin{table}[t]
\centering
\tablestyle{1.59pt}{1.2}\begin{tabular}{l|l|x{32}|ccc|c|c}
Method & Backbone & Scale & PQ & PQ$^{t}$ & PQ$^{s}$ & FPS$\uparrow$ & GPU\\ \shline
BGRNet~\cite{wu2020bidirectional} & R50-FPN & 800,1333 & 43.2 & 49.8 & 33.4 & - & - \\
K-Net~\cite{zhang2021k} & R50-FPN & 800,1333 & 47.1 & 51.7 & 40.3 & - & - \\
PanSegFormer~\cite{li2021panoptic} & R50 & 800,1333 & 49.6 & 54.4 & 42.4 & - & - \\
Max-DeepLab~\cite{wang2021max} & Max-S & 800,1333 & 48.4 & 53.0 & 41.5 & 7.6 & V100 \\
Mask2Former~\cite{cheng2021masked} & R50 & 800,1333 & 51.9 & 57.7 & 43.0 & 8.6 & V100 \\
UPSNet~\cite{xiong2019upsnet} & R50-FPN & 800,1333 & 42.5 & 48.5 & 33.4 & 9.1 & V100 \\
PanopticFCN~\cite{li2021fully} & R50-FPN & 800,1333 & 44.3 & 50.0 & 35.6 & 9.2 & V100 \\
LPSNet~\cite{hong2021lpsnet} & R50-FPN & 800,1333 & 39.1 & 43.9 & 30.1 & 9.3 & V100 \\
RealTimePan~\cite{hou2020real} & R50-FPN & 800,1333 & 37.1 & 41.0 & 30.7 & 15.9 & V100 \\
PanopticFPN~\cite{kirillov2019panoptic} & R50-FPN & 800,1333 & 41.5 & 48.3 & 31.2 & 17.5 & V100 \\
MaskFormer~\cite{cheng2021per} & R50 & 800,1333 & 46.5 & 51.0 & 39.8 & 17.6 & V100 \\
PanopticDeepLab~\cite{cheng2020panoptic} & R50 & 641,641 & 35.1 & - & - & 20.0 & V100 \\ \hline
\textbf{YOSO}, \emph{ours} & R50 & 800,1333 & 48.4 & 53.5 & 40.8 & 23.6 & V100 \\
\textbf{YOSO}, \emph{ours} & R50 & 512,800 & 46.4 & 50.7 & 40.0 & 45.6 & V100
\end{tabular} \vspace{-2.4mm}
\caption{Panoptic segmentation on the \textbf{COCO} \textit{validation} set.}\label{tab1}
\end{table}
\begin{table}[t]
\centering
\tablestyle{1.34pt}{1.2}\begin{tabular}{l|l|x{36}|ccc|c|c}
Method & Backbone & Scale & PQ & PQ$^{t}$ & PQ$^{s}$ & FPS$\uparrow$ & GPU \\ \shline
PanopticFPN~\cite{kirillov2019panoptic} & R50-FPN & 1024,2048 & 57.7 & 51.6 & 62.2 & - & - \\
Seamless~\cite{Porzi_2019_CVPR} & R50 & 1024, 2048 & 59.8 & 54.6 & 63.6 & - & - \\
PanopticFCN~\cite{li2021fully} & R50-FPN & 1024,2048 & 61.4 & 54.8 & 66.6 & - & -\\
Mask2Former~\cite{cheng2021masked} & R50 & 1024,2048 & 62.1 & 54.9 & 67.3 & 4.1 & V100 \\
UPSNet~\cite{xiong2019upsnet} & R50-FPN & 1024,2048 & 59.3 & 54.6 & 62.7 & 7.5 & V100 \\
LPSNet~\cite{hong2021lpsnet} & R50-FPN & 1024,2048 & 59.7 & 54.0 & 63.9 & 7.7 & V100 \\
PanopticDeepLab~\cite{cheng2020panoptic} & R50-FPN & 1024,2048 & 59.7 & - & - & 8.5 & V100 \\
FPSNet~\cite{de2020fast} & R50-FPN & 1024,2048 & 55.1 & - & - & 8.8 & Titan \\
RealTimePan~\cite{hou2020real} & R50-FPN & 1024,2048 & 58.8 & 52.1 & 63.7 & 10.1 & V100 \\
\hline 
\textbf{YOSO}, \emph{ours} & R50 & 1024,2048 & 59.7 & 51.0 & 66.1 & 11.1 & V100 \\
\textbf{YOSO}, \emph{ours} & R50 & 512,1024 & 52.5 & 43.5 & 59.1 & 22.6 & V100
\end{tabular} \vspace{-2.4mm}
\caption{Panoptic segmentation on the \textbf{Cityscapes} \textit{validation} set.}\label{tab2}
\end{table}
\begin{table}[t]
\centering
\tablestyle{1.97pt}{1.2}\begin{tabular}{l|l|x{32}|ccc|c|c}
Method & Backbone & Scale & PQ & PQ$^{t}$ & PQ$^{s}$ & FPS$\uparrow$ & GPU\\ \shline
BGRNet~\cite{wu2020bidirectional} & R50-FPN & - & 31.8 & 34.1 & 27.3 & - & - \\
PanSegFormer~\cite{li2021panoptic} & R50 & - & 36.4 & 35.3 & 38.6 & - & - \\
MaskFormer~\cite{cheng2021per} & R50 & 640,2560 & 34.7 & 32.2 & 39.7 & - & - \\ 
Mask2Former~\cite{cheng2021masked} & R50 & 640,2560 & 39.7 & 39.0 & 40.9 & 11.1 & V100 \\ \hline
\textbf{YOSO}, \emph{ours} & R50 & 640,2560 & 38.0 & 37.3 & 39.4 & 35.4 & V100
\end{tabular} \vspace{-2.4mm}
\caption{Panoptic segmentation on the \textbf{ADE20K} \textit{validation} set.}\label{tab3}
\end{table}
\begin{table}[t]
\centering
\tablestyle{1.46pt}{1.2}\begin{tabular}{l|l|x{34}|ccc|c|c}
Method & Backbone & Scale & PQ & PQ$^{t}$ & PQ$^{s}$ & FPS$\uparrow$ & GPU\\ \shline
AdaptIS~\cite{sofiiuk2019adaptis} & R50 & - & 32.0 & 39.1 & 26.6 & - & - \\
Seamless~\cite{Porzi_2019_CVPR} & R50 & - & 36.2 & 33.6 & 40.0 & - & - \\
LPSNet~\cite{hong2021lpsnet} & R50-FPN & - & 36.5 & 33.2 & 41.0 & - & - \\ 
PanopticFCN~\cite{li2021fully} & R50-FPN & - & 36.9 & 32.9 & 42.3 & - & - \\ 
PanopticDeepLab~\cite{cheng2020panoptic} & R50 & 2176,2176 & 33.3 & - & - & 3.5 & V100 \\
Mask2Former~\cite{cheng2021masked} & R50 & 2048,2048 & 36.3 & - & - & 3.2 & A100 \\  \hline
\textbf{YOSO}, \emph{ours} & R50 & 2048,2048 & 34.1 & 24.3 &  47.2 & 7.1 & A100
\end{tabular} \vspace{-2.4mm}
\caption{Panoptic segmentation on the \textbf{Mapillary} \textit{validation} set.}\label{tab4}
\end{table}
\begin{figure}
\centering
\includegraphics[width=1.0\linewidth]{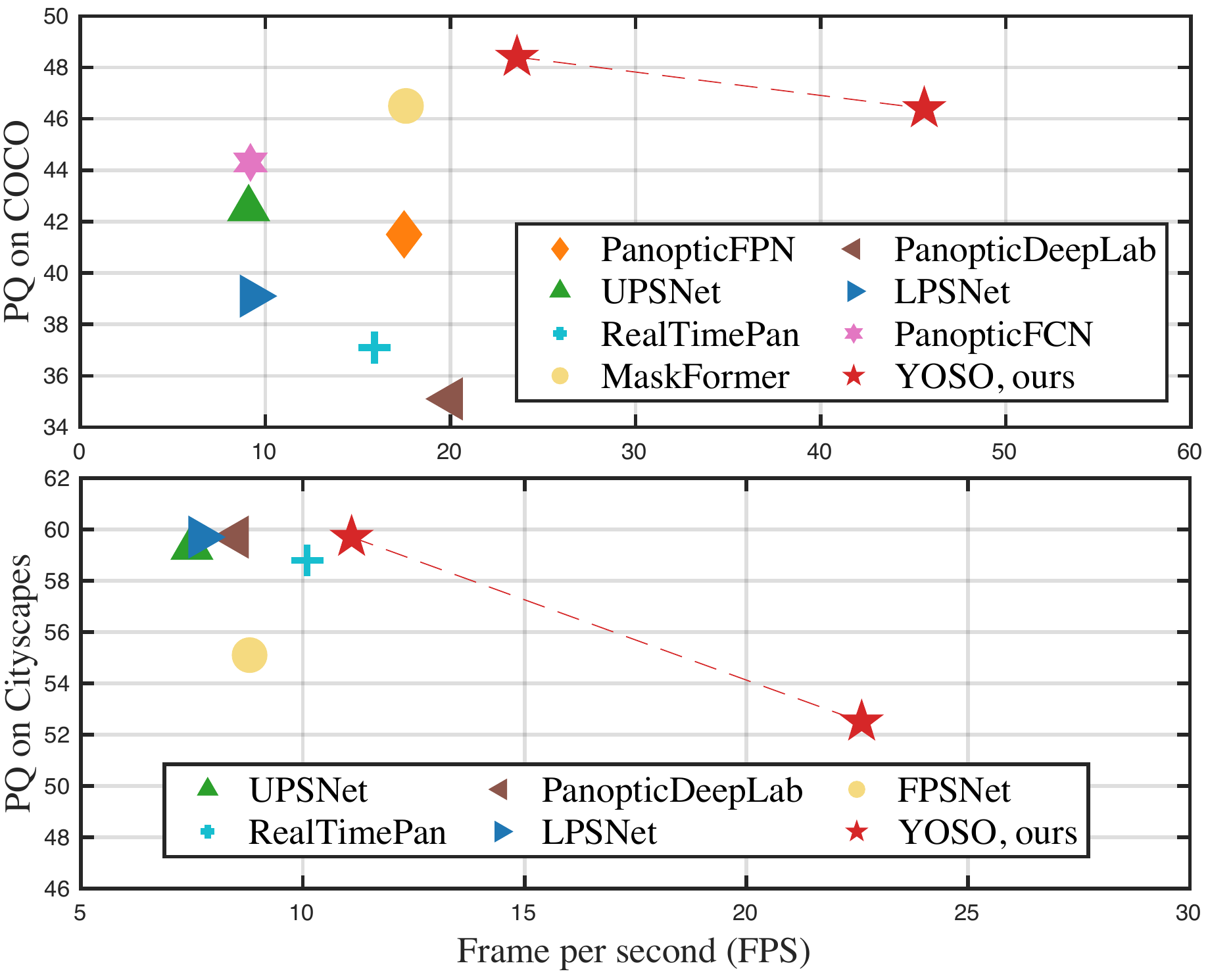}\vspace{-2.4mm}
\caption{
\textbf{FPS \wrt PQ} on the COCO and Cityscapes datasets.
}\label{fig4}
\end{figure}
\subsection{Implementation Details}
For the COCO dataset, we used a batch size of 16 and set the learning rate to 0.0001.
The models were trained for 370k iterations with large-scale jitter augmentation~\cite{ghiasi2021simple}.
For the Cityscapes and Mapillary datasets, we trained the models with a batch size of 16, the learning rate set to 0.0001, and the training schedule set to 180k iterations.
For the ADE20K dataset, we set the batch size to 16, the learning rate to 0.0001, and the models were trained for 30k iterations.
The ResNet50~\cite{he2016deep} (denoted as R50) pre-trained on the ImageNet~\cite{deng2009imagenet} dataset is employed as our backbone for the four datasets.
We used ResNet50~\cite{he2016deep} (denoted as R50) pre-trained on ImageNet~\cite{deng2009imagenet} as the backbone for all four datasets, and set the hidden dimension $d$ to 256 through experimentation.
Since the COCO dataset contains images from various scenes, ranging from indoor to outdoor, we conducted ablation studies on this dataset.
In the ablation studies, the models were trained with 270k iterations.

\subsection{Main Results}
The results of panoptic segmentation on the COCO dataset are presented in Tab.~\ref{tab1}.
We have the following findings.
First, YOSO is significantly faster than predominant efficient panoptic segmentation models such as PanopticFPN~\cite{kirillov2019panoptic} and RealTimePan~\cite{hou2020real}.
Specifically, YOSO achieves a PQ of 48.4 and an FPS of 23.6 with an input image scale of (800, 1333).
This PQ is 11.3 higher than that of RealTimePan, while the speed is approximately 1.5$\times$ faster.
Furthermore, when scaling the input image to (512, 800), YOSO is around 2.3$\times$ faster than the previous fastest model, PanopticDeepLab, while achieving an approximately 11.0-point higher PQ.
Second, YOSO achieves comparable accuracy with state-of-the-art models such as MaskFormer~\cite{cheng2021per}, Mask2Former~\cite{cheng2021masked}, Max-DeepLab~\cite{wang2021max}, and K-Net~\cite{zhang2021k}.
For example, YOSO outperforms MaskFormer and K-Net by 1.9 and 1.3 PQ, respectively, and achieves the same PQ performance as Max-DeepLab.
Although the PQ of YOSO is 3.5 points lower than that of Mask2Former, YOSO is 2.7$\times$ faster than Mask2Former.

The results of panoptic segmentation on the Cityscapes dataset are presented in Tab.~\ref{tab2}.
YOSO is the fastest model with competitive accuracy among state-of-the-art approaches.
For example, YOSO achieves 59.7 PQ and 11.1 FPS with the input image size of (1024, 2048), which is 4.7 points higher than FPSNet.
Moreover, when reducing the input image scale to (512, 1024), YOSO achieves an accuracy of 52.5 PQ with 22.6 FPS.
%

In Tab.~\ref{tab3} and Tab.~\ref{tab4}, we show the panoptic segmentation results on the ADE20K and the Mapillary Vistas datasets, respectively, to evaluate the model generalization of YOSO.
On the ADE20K dataset, YOSO outperforms most previous methods such as PanSegFormer~\cite{li2021panoptic} and MaskFormer~\cite{cheng2021per} in terms of both speed and accuracy.
On the Mapillary Vistas dataset, although YOSO has a good PQ$^s$, the performance of PQ$^t$ lags behind that of state-of-the-art models.
This suggests that YOSO still has the potential to be improved on the Mapillary Vistas dataset.

Additionally, we plot the PQ \emph{w.r.t.} FPS results on the COCO and the Cityscapes datasets in Fig.~\ref{fig4}, which shows that YOSO runs faster and achieves competitive accuracy among state-of-the-art models.
In summary, the main results on the four datasets validate the good generalization and well-balanced speed-accuracy of YOSO.
\begin{table}[t]
\centering
\tablestyle{2.41pt}{1.2}\begin{tabular}{c|ccc|cc|c|c|c}
Aggregator & PQ & SQ & RQ & PQ$^{t}$ & PQ$^{s}$ & FLOPs & Latency ($\mu$s)$\downarrow$ & FPS$\uparrow$ \\ \shline
IFA & 47.5 & 82.2 & 56.9 & 52.7 & 39.7 & 16.6G & 4871$\pm$11 & 23.3 \\
CFA & 47.0 & 81.4 & 56.2 & 52.3 & 39.0 & 2.1G & 1877$\pm$52 & 29.2 \\ 
\end{tabular} \vspace{-2.4mm}
\caption{\textbf{Comparison of different aggregators.} The FLOPs and GPU latency were obtained from single modules with the setting of $d$=256, $c_2$=128, $c_3$=256, $c_4$=512, $c_5$=1024, and $h$=$w$=256.}\label{tab5}
\end{table}
\begin{table}[t]
\centering
\tablestyle{2.74pt}{1.2}\begin{tabular}{c|ccc|cc|c|c|c}
Attention & PQ & SQ & RQ & PQ$^{t}$ & PQ$^{s}$ & FLOPs & Latency ($\mu$s)$\downarrow$ & FPS$\uparrow$ \\ \shline
MHCA & 46.0 & 81.9 & 55.1 & 51.5 & 37.7 & 31.5M & 2608$\pm$210 & 27.3 \\ \hline
SDCA & 47.0 & 81.4 & 56.2 & 52.3 & 39.0 & 5.4M & 2183$\pm$279 & 29.2 \\ 
DCA & 46.9 & 82.0 & 55.8 & 51.9 & 38.7 & 15.5M & 1701$\pm$186 & 30.0 \\
PDCA & 43.7 & 81.3 & 52.3 & 49.3 & 35.3 & 5.2M & 1450$\pm$183 & 30.2 \\
DDCA & 46.6 & 82.3 & 55.9 & 52.3 & 38.4 & 0.3M & 1242$\pm$101 & 30.3
\end{tabular} \vspace{-2.4mm}
\caption{\textbf{Comparison of different attention modules.} The FLOPs and GPU latency were obtained from single modules with the setting of $n$=100, $d$=256, and $t$=3.
The modules tested included MHCA (Multi-Head Cross-Attention), DCA (Dynamic Convolution Attention), SDCA (Separable Dynamic Convolution Attention), PDCA (Pointwise Dynamic Convolution Attention), and DDCA (Depthwise Dynamic Convolution Attention).}\label{tab6}
\end{table}

\subsection{Ablation Study}
In order to examine the impact of different components on the speed and accuracy of YOSO, we conducted several ablation studies focusing on the feature pyramid aggregator and the separable dynamic decoder.
Specifically, we evaluated the effectiveness of the aggregation modules and the attention modules, which resulted in several interesting findings.
Moreover, we analyzed how variations in the number of attention blocks, kernel size, iteration stages, and proposal kernels affected the performance.
The details of our investigations are discussed below.

\textbf{Comparison of different aggregators.}
The results of using different aggregators are presented in Tab.~\ref{tab5}.
Specifically, we trained YOSO with IFA and CFA, respectively.
The PQ results indicate that IFA achieves higher accuracy than CFA, with PQ values of 47.5 and 47.0, respectively.
However, IFA has much larger FLOPs than CFA, with 16.6G compared to 2.1G, and a slower GPU latency, with 4871$\mu$s compared to 1877$\mu$s.
Considering that the learned parameters of IFA can be directly re-parameterized to CFA, we can train YOSO with IFA and infer with the re-parameterized CFA for better speed and accuracy.
\begin{figure}
\centering
\includegraphics[width=1.0\linewidth]{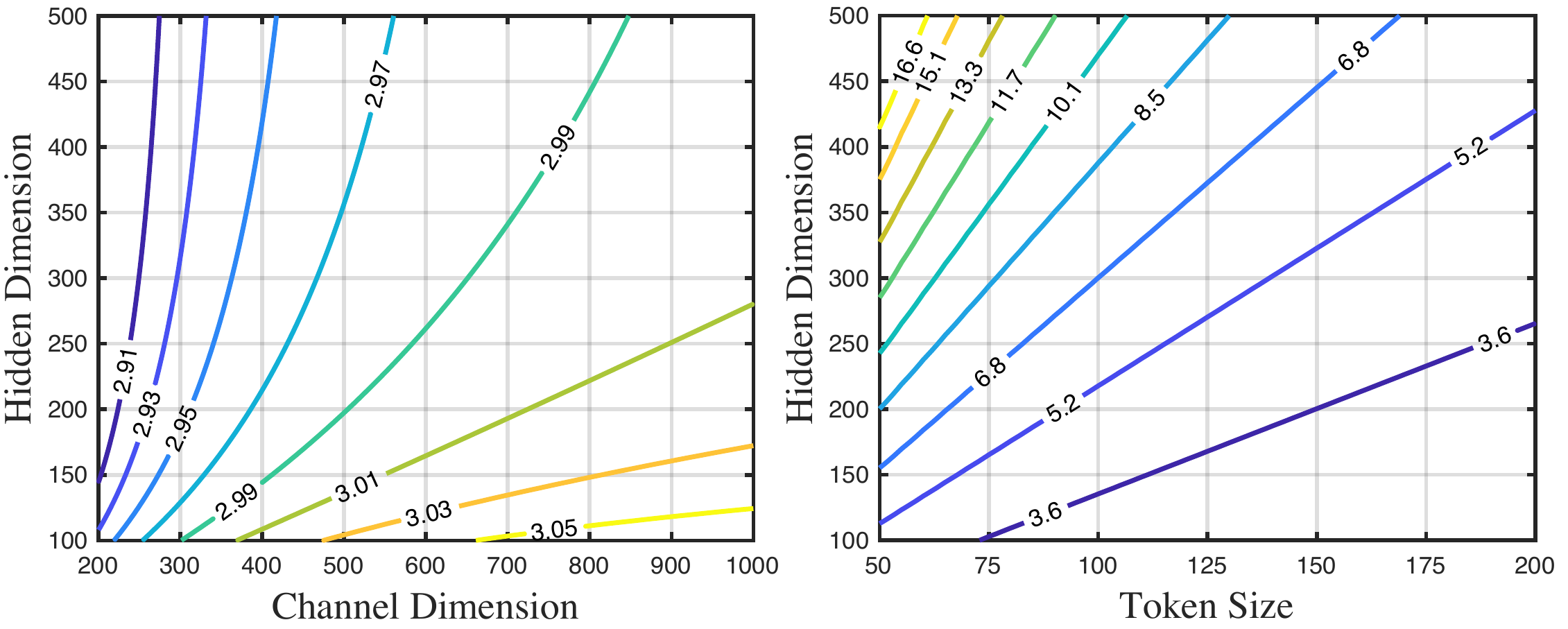}\vspace{-2.4mm}
\caption{
\textbf{Contours of FLOPs reduction ratio.} Left: CFA. Right: SDCA. The channel dimension of input feature maps for CFA and the hidden dimension of input tokens for SDCA are the key factors to reducing the number of FLOPs.
}\label{fig5}
\end{figure}

In Fig.~\ref{fig5} (left), we investigate the reduction ratio of FLOPs between CFA and IFA.
Specifically, we set the channel dimensions of the input feature maps $c_5$, $c_4$, $c_3$, and $c_2$ to be equal to $c$, and analyze how the input channel dimension $c$ and the output hidden dimension $d$ affect the FLOPs reduction ratio in Eq.~\ref{eq2}.
Our results indicate that increasing the input channel dimension and decreasing the output hidden dimension lead to an increase in the reduction ratio.
This implies that CFA will be more efficient when the dimension of the input channel is large.

\textbf{Comparison of different attention modules.}
We present a comparison of the effectiveness of different attention modules in Tab.~\ref{tab6}.
In terms of PQ performance, we have made two interesting findings.
First, we were surprised to find that the MHCA module did not perform better than DCA and SDCA.
The reason for this may be the difference between the basic operations in these two types of attention modules: DCA densely splits the hidden dimension into $d$ groups, while MHCA sparsely splits it into $t$ groups.
Second, PDCA showed inferior performance compared to the other modules, with a PQ of 43.7.
This is likely due to the fact that PDCA is the only module that does not apply cross-dimension interaction, which implies that interactions between hidden dimensions are significant for the attention module.
This observation is supported by the performance of DDCA, which only performs cross-token interaction and achieved a PQ of 46.7.
These two results suggest that the cross-dimension interaction may be more significant than the cross-token interaction in panoptic kernel generation, as the panoptic kernels are expected to be independent to represent different \emph{stuff} or \emph{things}.
In terms of FLOPs, we observed that the attention modules based on dynamic convolution require fewer FLOPs.
Specifically, DDCA exhibits the lowest computational cost, requiring only 0.3M FLOPs.
%
%
%
In terms of GPU latency, we made two interesting observations.
First, although the time complexity of MHCA, DCA, SDCA, and PDCA are all $O(n^2d)$, the speedup on the GPU latency is remarkable.
Second, we found that SDCA runs slower than DCA, contrary to what the FLOPs analysis suggests.
We speculate that the additional convolutional and fully connected layers in SDCA are executed serially rather than in parallel, which may result in longer execution times in practice.

Fig.~\ref{fig5} (right) presents the analysis of the FLOPs reduction ratio, as given by Eq.~\ref{eq10}.
We fix the kernel size $t$ to 3, and investigate the effect of the token size $n$ and the hidden dimension $d$ on the reduction ratio.
The results indicate that the reduction ratio enlarges with the token size decreases and the hidden dimension increases.
Specifically, DCA performs better when the token size is smaller than the hidden dimension, indicating its suitability for vision tasks where the token size is smaller than the hidden dimension.

\begin{table}[t]
\centering
\tablestyle{2.91pt}{1.2}\begin{tabular}{c|ccc|ccc|ccc|c}
\#Blocks & PQ & SQ & RQ & PQ$^{t}$ & SQ$^{t}$ & RQ$^{t}$ & PQ$^{s}$ & SQ$^{s}$ & RQ$^{s}$ & FPS$\uparrow$ \\ \shline
$N$=1 & 46.4 & 81.7 & 55.7 & 52.1 & 82.3 & 62.1 & 37.9 & 79.9 & 46.0 & 30.1 \\
$N$=2 & 47.0 & 81.4 & 56.2 & 52.3 & 83.1 & 62.2 & 39.0 & 78.8 & 47.1 & 29.2 \\ 
$N$=3 & 47.5 & 82.1 & 56.8 & 52.6 & 83.2 & 62.6 & 39.7 & 80.4 & 48.3 & 28.9 \\
$N$=4 & 47.2 & 81.8 & 56.7 & 52.5 & 82.9 & 62.6 & 39.3 & 80.1 & 47.6 & 28.7 
\end{tabular} \vspace{-2.4mm}
\caption{\textbf{Influence of attention block number.} The PQ accuracy increases with more blocks, while the FPS decreases.}\label{tab7}
\end{table}
\begin{table}[t]
\centering
\tablestyle{2.41pt}{1.2}\begin{tabular}{c|ccc|ccc|ccc|c}
Kernel Size & PQ & SQ & RQ & PQ$^{t}$ & SQ$^{t}$ & RQ$^{t}$ & PQ$^{s}$ & SQ$^{s}$ & RQ$^{s}$ & FPS$\uparrow$ \\ \shline
$t$=1 & 46.8 & 82.3 & 55.9 & 52.4 & 83.4 & 62.3 & 38.5 & 80.6 & 46.4 & 30.6 \\
$t$=3 & 47.0 & 81.4 & 56.2 & 52.3 & 83.1 & 62.2 & 39.0 & 78.8 & 47.1 & 29.2 \\ 
$t$=5 & 47.3 & 81.3 & 56.8 & 52.2 & 82.0 & 62.2 & 40.0 & 80.3 & 48.6 & 28.3 \\
$t$=7 & 47.1 & 81.2 & 56.3 & 52.4 & 83.2 & 62.3 & 39.8 & 80.2 & 48.4 & 27.6 
\end{tabular} \vspace{-2.4mm}
\caption{\textbf{Results of using different convolution kernel sizes.} The PQ performance saturates when $t$=5.}\label{tab8}
\end{table}

\textbf{Number of attention blocks.}
To assess the effectiveness of the separable dynamic convolution attention in the separable dynamic decoder, we varied the number of blocks and analyzed the speed-accuracy trade-off, as presented in Tab.~\ref{tab7}.
The results show that the PQ accuracy improves with additional blocks, but at the expense of lower FPS.
Consequently, we selected $N=2$ as a compromise between speed and accuracy for YOSO.

\textbf{Kernel size of dynamic convolution.}
Tab.~\ref{tab8} shows the effectiveness of different kernel sizes for the separable dynamic convolution attention module.
%
%
The PQ performance confirms our observation in Tab.~\ref{tab6}, suggesting that the cross-dimension interaction plays a significant role in the attention modules.
Specifically, reducing the kernel size from 5 to 1 leads to a degradation in the PQ performance from 47.3 to 46.8.
Furthermore, the performance reaches saturation when the kernel size is increased from 5 to 7.
\textbf{Iteration stages.}
To assess the impact of the number of stages, we conducted experiments with different numbers of stages and report the results in Tab.~\ref{tab9}.
Our results show that increasing the number of stages improves the PQ performance, but it comes at the cost of decreasing the FPS performance.
We observed that the trade-off between speed and accuracy is best achieved when using $T=2$ stages. Hence, we selected this configuration for YOSO.

\textbf{Number of proposal kernels.}
We investigate the impact of the number of proposal kernels in Tab.~\ref{tab10}.
The results indicate that the PQ performance improves when increasing the number of proposal kernels from 50 to 100, and saturates at 150.
Meanwhile, the speed decreases when increasing the number of proposal kernels.
The setting of $n$=100 well balances the accuracy and speed for YOSO.
\begin{table}[t]
\centering
\tablestyle{2.97pt}{1.2}\begin{tabular}{c|ccc|ccc|ccc|c}
\#Stages & PQ & SQ & RQ & PQ$^{t}$ & SQ$^{t}$ & RQ$^{t}$ & PQ$^{s}$ & SQ$^{s}$ & RQ$^{s}$ & FPS$\uparrow$ \\ \shline
$T$=1 & 45.7 & 80.4 & 53.6 & 50.3 & 82.1 & 59.6 & 36.5 & 77.3 & 45.2 & 30.1 \\
$T$=2 & 47.0 & 81.4 & 56.2 & 52.3 & 83.1 & 62.2 & 39.0 & 78.8 & 47.1 & 29.2 \\ 
$T$=3 & 47.5 & 81.9 & 56.9 & 52.8 & 82.9 & 63.0 & 39.4 & 80.4 & 47.7 & 28.9 
\end{tabular} \vspace{-2.4mm}
\caption{\textbf{Impact of iteration stages.} The PQ accuracy increases while the FPS decreases with more iteration stages.}\label{tab9}
\end{table}
\begin{table}[t]
\centering
\tablestyle{2.51pt}{1.2}\begin{tabular}{c|ccc|ccc|ccc|c}
\#Proposals & PQ & SQ & RQ & PQ$^{t}$ & SQ$^{t}$ & RQ$^{t}$ & PQ$^{s}$ & SQ$^{s}$ & RQ$^{s}$ & FPS$\uparrow$ \\ \shline
$n$=50 & 44.2 & 80.1 & 52.6 & 49.8 & 81.2 & 59.2 & 36.1 & 78.4 & 44.1 & 33.7 \\
$n$=100 & 47.0 & 81.4 & 56.2 & 52.3 & 83.1 & 62.2 & 39.0 & 78.8 & 47.1 & 29.2 \\ 
$n$=150 & 46.9 & 81.0 & 55.7 & 52.0 & 82.7 & 61.5 & 38.9 & 78.6 & 46.8 & 26.3 \\
$n$=200 & 46.7 & 81.1 & 55.2 & 51.5 & 82.5 & 61.2 & 38.7 & 78.5 & 46.2 & 24.3
\end{tabular} \vspace{-2.4mm}
\caption{\textbf{Different number of proposals.} The PQ accuracy increases with more proposals and saturates when $n$=150.}\label{tab10}
\end{table}

\section{Conclusion}
In this paper, we propose a real-time panoptic segmentation framework, termed YOSO.
With YOSO, \emph{you only need to segment once} for the masks of both foreground \emph{things} and background \emph{stuff}.
YOSO includes a feature pyramid aggregator and a separable dynamic decoder to accelerate the pipeline.
The CFA module in the feature pyramid aggregator re-parameters the IFA module, reducing FLOPs without extra costs.
The SDC module in the separable dynamic decoder performs weight-sharing multi-head cross-attention, enhancing both speed and accuracy.
Our extensive experiments demonstrate that YOSO is significantly faster than other prominent panoptic segmentation methods while maintaining competitive PQ performance.
Given its effectiveness and simplicity, we hope YOSO can serve as a strong baseline and bring fresh insights for future research on real-time panoptic segmentation.

\vspace{-4mm}\paragraph{Acknowledgements.}
\footnotesize{This work was supported by National Key R\&D Program of China (2022ZD0118202), National Science Fund for Distinguished Young Scholars (No.62025603), National Natural Science Foundation of China (No.U21B2037, No.U22B2051, No.62176222, No.62176223, No.62176226, No.62072386, No.62072387, No.62072389, No.62002305 and No.62272401), and Natural Science Foundation of Fujian Province of China (No.2021J01002, No.2022J06001).}

{\small
\bibliographystyle{ieee_fullname}
\bibliography{egbib}
}


\begin{appendices}
\onecolumn

\section{Qualitative Results for Panoptic Segmentation} 
%
%
%
%
%
\begin{figure*}[h]
\centering
\includegraphics[width=1.0\linewidth]{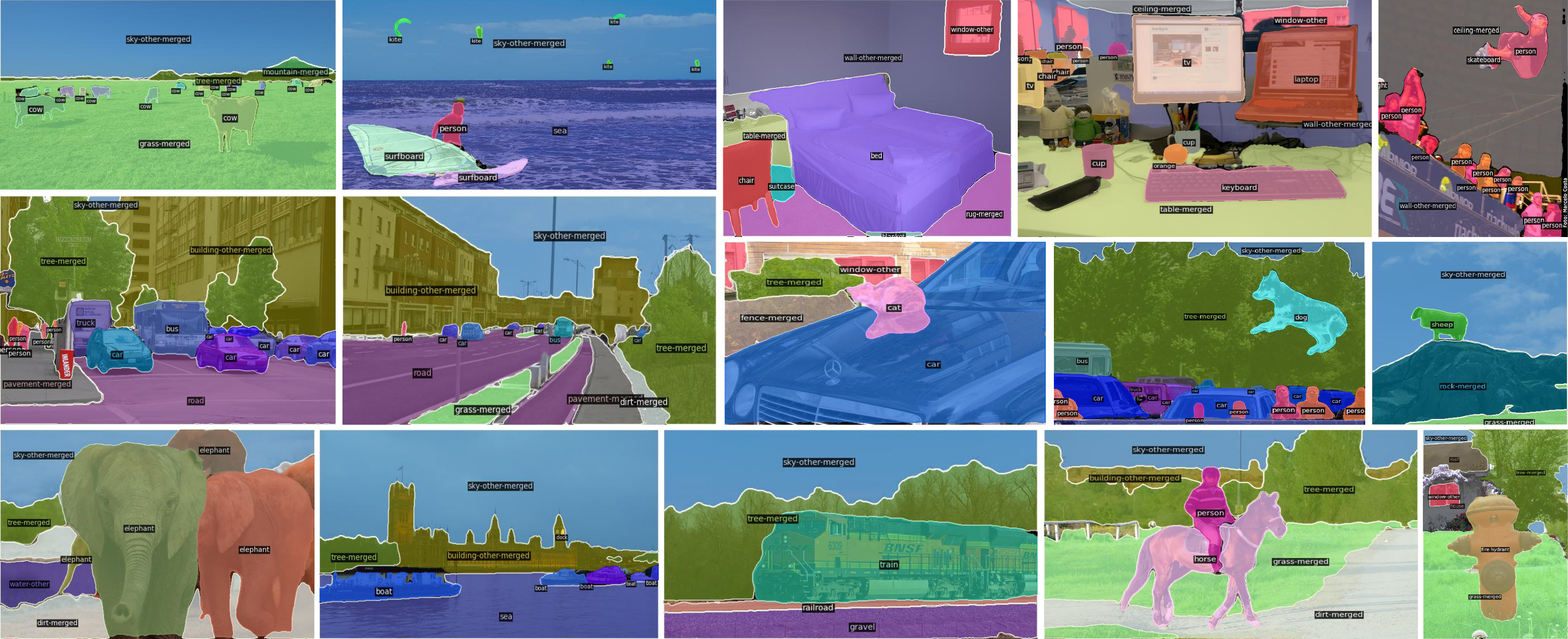} 
\caption{
Panoptic segmentation on the \textbf{COCO} \textit{validation} set.
}\label{fig6}
\end{figure*}
\begin{figure*}[h]
\centering
\includegraphics[width=1.0\linewidth]{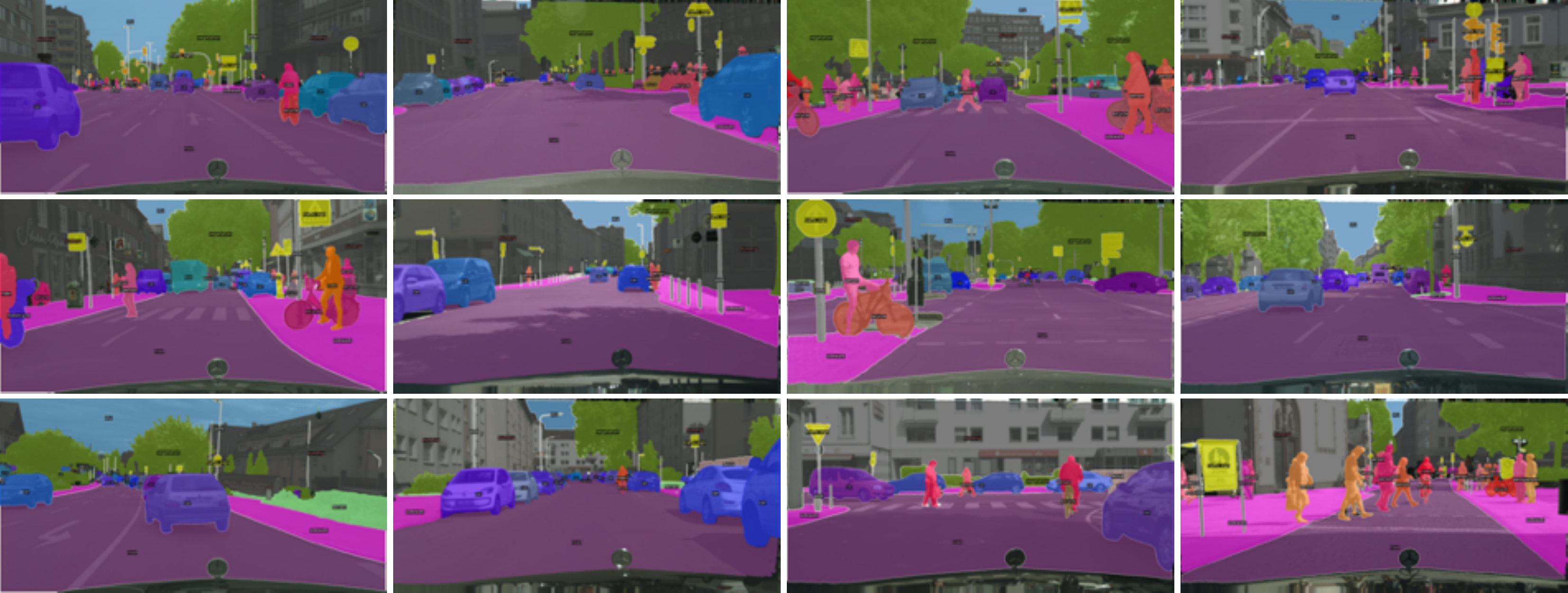} 
\caption{
Panoptic segmentation on the \textbf{Cityscapes} \textit{validation} set.
}\label{fig7}
\end{figure*}
\begin{figure*}[h]
\centering
\includegraphics[width=1.0\linewidth]{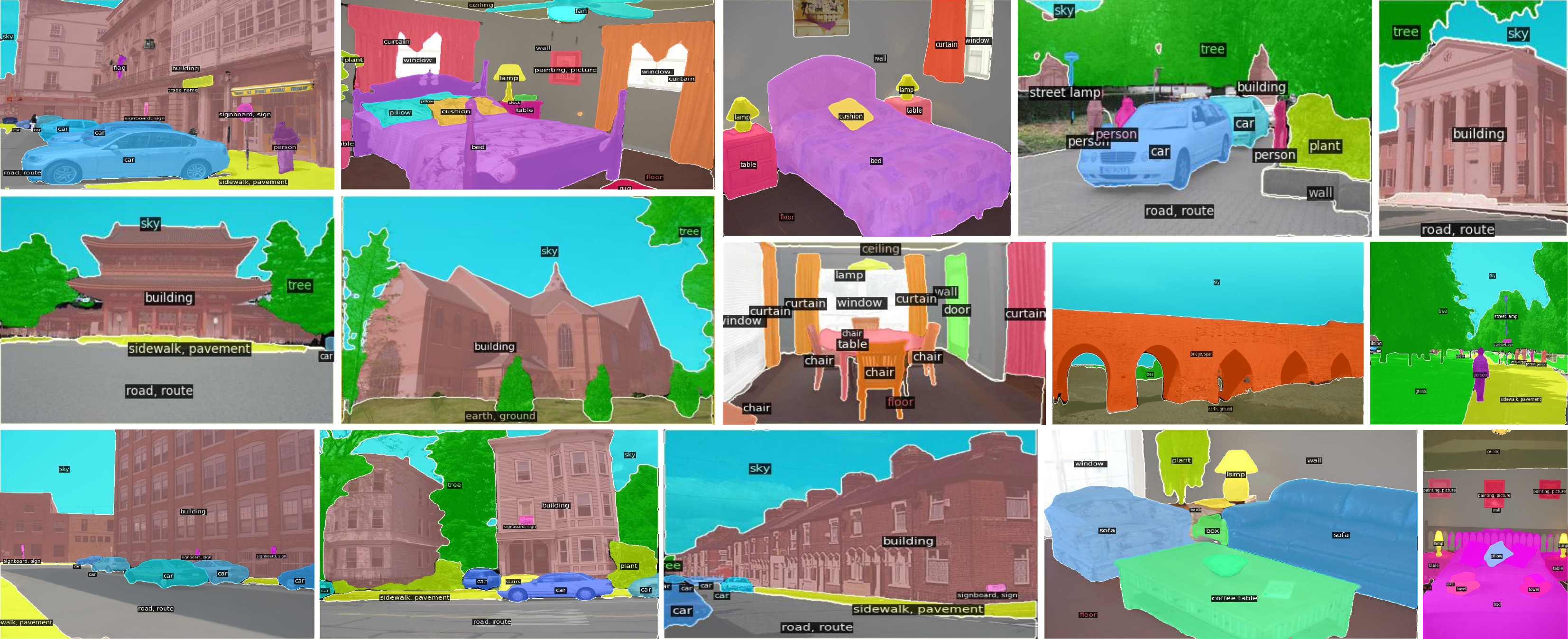} 
\caption{
Panoptic segmentation on the \textbf{ADE20K} \textit{validation} set.
}\label{fig8}
\end{figure*}
\begin{figure*}[h]
\centering
\includegraphics[width=1.0\linewidth]{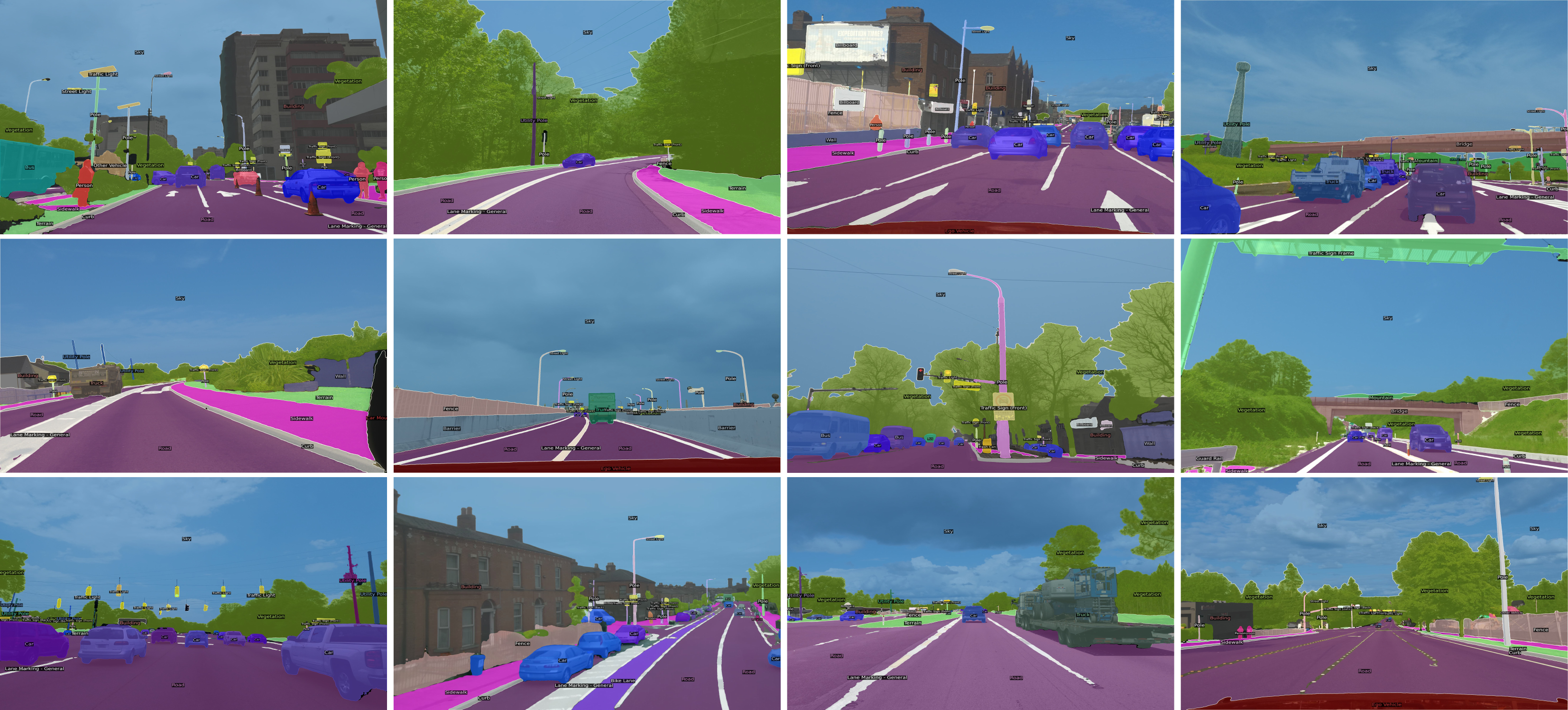} 
\caption{
Panoptic segmentation on the \textbf{Mapillary Vistas} \textit{validation} set.
}\label{fig9}
\end{figure*}
%
\section{Quantitive Results on Real-Time Instance Segmentation}
Solving the instance segmentation task efficiently is one of the keys to achieving real-time panoptic segmentation.
Therefore, we study the performance of YOSO for real-time instance segmentation in Tab.~\ref{tab11}.
Note that the model is not specifically trained for instance segmentation.
We show the results from the model trained on the COCO training set for panoptic segmentation.
From the results shown in Tab.~\ref{tab11}, we find that YOSO also achieves competitive performance for instance segmentation on the COCO validation set.
When scaling the input images to 550, YOSO achieves 38.7 FPS and 34.7 mAP.
The speed is only 5.9 FPS lower than the current state-of-the-art model, \ie, SparseInst, while the mAP is 0.3 points higher.
Specifically, the performance of YOSO on large objects, \ie, AP$^l$, is better than all the state-of-the-art models.
For example, when scaling the input images to 448, YOSO still achieves 57.6 AP$^l$ for large objects, which is approximately 2.2 points higher than the performance of SOLOv2.
The result suggests that YOSO is good at segmenting large objects for instance segmentation.
\begin{table}[t]
\centering
\tablestyle{4.6pt}{1.2}\begin{tabular}{l|l|c|ccc|ccc|c|c}
Method & Backbone & Scale & AP & AP$_{50}$ & AP$_{75}$ & AP$^{s}$ & AP$^{m}$ & AP$^{l}$ & GPU & FPS \\ \shline
YOLACT~\cite{bolya2019yolact} & DarkNet-53 & 550 & 28.9 & 46.9 & 30.3 & 9.8 & 30.9 & 47.3 & 2080Ti & 45.9 \\
YOLACT++~\cite{bolya2020yolact++} & ResNet-50 & 550 & 33.7 & 52.7 & 35.5 & 11.9 & 36.6 & 54.6 & 2080Ti & 40.8 \\
BlendMask~\cite{chen2020blendmask} & ResNet-50 & 550 & 34.5 & 54.7 & 36.5 & 14.4 & 37.7 & 52.1 & 2080Ti & 35.6 \\
SOLOv2~\cite{wang2020solov2} & ResNet-50 & 448 & 33.7 & 53.3 & 35.6 & 11.3 & 36.9 & 55.4 & 2080Ti & 39.6 \\
OrienMask~\cite{du2021real} & DarkNet-53 & 544 & 34.5 & 56.0 & 35.8 & 16.8 & 38.5 & 49.1 & 2080Ti & 41.9 \\
SparseInst~\cite{cheng2022sparse} & ResNet-50 & 608 & 34.4 & 55.2 & 36.1 & 14.2 & 36.8 & 51.9 & 2080Ti & 44.6 \\ \hline
\textbf{YOSO}, \emph{ours} & ResNet-50 & 448 & 33.0 & 52.8 & 34.6 & 11.3 & 35.4 & 57.6 & 2080Ti & 46.1 \\
\textbf{YOSO}, \emph{ours} & ResNet-50 & 550 & 34.7 & 55.0 & 36.4 & 13.2 & 37.6 & 58.6 & 2080Ti & 38.7 \\
\textbf{YOSO}, \emph{ours} & ResNet-50 & 608 & 35.6 & 56.3 & 37.5 & 14.4 & 39.0 & 59.2 & 2080Ti & 33.5 \\
\end{tabular} \vspace{-2.4mm}
\caption{Real-time instance segmentation results on the \textbf{COCO} \textit{validation} set.\label{tab11}}
\end{table}

\end{appendices}
\end{document}